\documentclass[conference]{IEEEtran}
\IEEEoverridecommandlockouts
\usepackage{cite}
\usepackage{amsmath,amssymb,amsfonts}
\usepackage{algorithmic}
\usepackage{graphicx}
\usepackage{textcomp}
\usepackage{xcolor}
\usepackage{hyperref}
\usepackage{cleveref}
\usepackage{colortbl}
\usepackage{xcolor}
\usepackage{booktabs}
\usepackage{adjustbox}
\usepackage{caption}
\usepackage{csquotes}
\usepackage{multirow}
\definecolor{tablegray}{HTML}{D8D8D8}
\definecolor{tableup}{RGB}{191,121,64}
\definecolor{tabledown}{HTML}{D8D8D8}
\def\BibTeX{{\rm B\kern-.05em{\sc i\kern-.025em b}\kern-.08em
    T\kern-.1667em\lower.7ex\hbox{E}\kern-.125emX}}
    
\begin{document}

\title{FRFDet: Efficient UAV Small Object Detection with Symmetric Sampling and Scalable Fusion}

% \author{\IEEEauthorblockN{1\textsuperscript{st} Yunzhong Si}
% \IEEEauthorblockA{\textit{College of Computer Science and Technology} \\
% \textit{Zhejiang Normal University}\\
% Jinhua, China \\
% siyunzhong@zjnu.edu.cn}
% \and
% \IEEEauthorblockN{2\textsuperscript{nd} Xu Huiying}
% \textit{Zhejiang Normal University}\\
% Jinhua, China \\
% xhy@zjnu.edu.cn}
% \and
% \IEEEauthorblockN{3\textsuperscript{rd} Xinzhong Zhu}
% \textit{Zhejiang Normal University}\\
% Jinhua, China \\
% zxz@zjnu.edu.cn}
% \and
% \IEEEauthorblockN{4\textsuperscript{th} Yang Liu}
% \textit{Zhejiang Normal University}\\
% Jinhua, China \\
% liuyang@zjnu.edu.cn}
% \and
% \IEEEauthorblockN{5\textsuperscript{th} Yao Dong}
% \IEEEauthorblockA{\textit{College of Computer Science and Technology} \\
% \textit{Zhejiang Normal University}\\
% Jinhua, China \\
% dongyao@zjnu.edu.cn}
% \and
% \IEEEauthorblockN{6\textsuperscript{th} Wenhao Zhang}
% \IEEEauthorblockA{\textit{College of Computer Science and Technology} \\
% \textit{Zhejiang Normal University}\\
% Jinhua, China \\
% zwh2012918201@zjnu.edu.cn}
% \and
% Beijing, China \\
% jason.li@geekplus.com}
% }

\author{
\IEEEauthorblockN{
Yunzhong Si$^{1}$, Huiying Xu$^{1}$, Xinzhong Zhu$^{1}$, Yang Liu$^{2}$, Yao Dong$^{1}$, Wenhao Zhang$^{1}$, Hongbo Li$^{3}$ \\
}
\IEEEauthorblockA{
$^{1}$College of Computer Science and Technology, Zhejiang Normal University, Jinhua, China \\
$^{2}$Hangzhou School of Automation, Zhejiang Normal University, Hangzhou, China \\
$^{3}$Beijing Geekplus Technology Co., Ltd, Beijing, China \\
\{siyunzhong, xhy, zxz, liuyang, dongyao, zwh2012918201\}@zjnu.edu.cn, jason.li@geekplus.com
}
}

% \author{
% Yunzhong Si$^{1}$, Huiying Xu$^{1}$, Xinzhong Zhu$^{1}$, Yang Liu$^{2}$, Yao Dong$^{1}$, Wenhao Zhang$^{1}$, Hongbo Li$^{3}$ \\
% \small
% $^{1}$College of Computer Science and Technology, Zhejiang Normal University, Jinhua, China \\
% $^{2}$Hangzhou School of Automation, Zhejiang Normal University, Hangzhou, China \\
% $^{3}$Beijing Geekplus Technology Co., Ltd, Beijing, China \\
% \small
% \{siyunzhong, xhy, zxz, liuyang, dongyao, zwh2012918201\}@zjnu.edu.cn, jason.li@geekplus.com
% }

\maketitle

\begin{abstract}
Small object detection in Unmanned Aerial Vehicle (UAV) imagery remains challenging under adverse conditions, including complex weather, low illumination, and sensor noise. These challenges mainly stem from severe background clutter, fine-grained detail degradation, and suboptimal semantic–spatial feature fusion, which jointly hinder robust small-object representation. To this end, we propose FRFDet, a lightweight yet effective single-stage detector tailored for UAV-based small object detection. FRFDet proposes two plug-and-play modules: Inverse Bidirectional Sampling (IBS) and Scale-Feature Relationship Cross-Fusion (SFRCF). IBS preserves critical spatial details via channel expansion–compression and bidirectional pattern reconstruction, improving feature alignment. SFRCF explicitly models scale-dependent fusion behaviors, revealing that inter-group element-wise multiplication favors compact models, while inter-group additive fusion benefits larger architectures. Extensive experiments on VisDrone, UAVDT, HazyDet, and MS COCO demonstrate that FRFDet achieves state-of-the-art performance among lightweight detectors with low computational cost, compact parameters, and fast inference, making it well suited for resource-constrained UAV platforms. The source code is available at: \url{https://github.com/HZAI-ZJNU/FRFDet}
\end{abstract}


\section{Introduction}

\label{sec:intro}

\begin{figure}[t]
    \centering
    \includegraphics[width=1\linewidth]{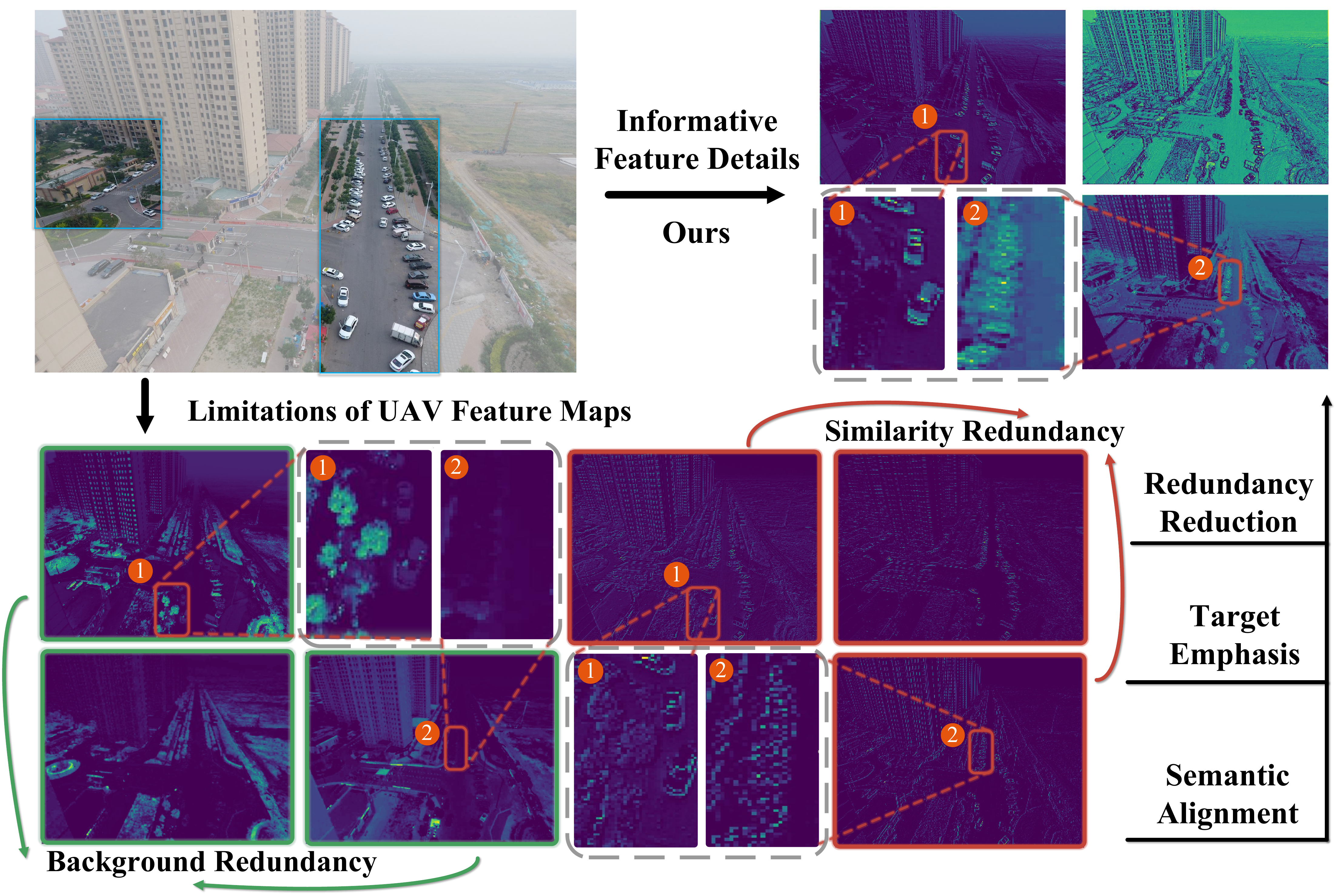}
    \caption{Feature map comparison. Standard sampling shows strong redundancy, while our method enhances small-object saliency and semantic consistency.}
    \label{fig:ReduancyInVisDrone}
\end{figure}

With the rapid evolution of deep detection paradigms, including DCNN-based frameworks \cite{FasterRCNN, YOLOv7}, Vision Transformers \cite{ViT, Swin, RTDETR}, and recent sequence models such as Mamba \cite{Mamba}, generic object detection has achieved remarkable performance on benchmarks such as MS COCO \cite{MSCOCO}. Nevertheless, small object detection in Unmanned Aerial Vehicle (UAV) imagery remains an open problem. This difficulty mainly arises from two intrinsic factors: (1) information sparsity, where objects smaller than $32 \times 32$ pixels are progressively degraded by downsampling operations \cite{SoftPool}, causing low-frequency background features to dominate; and (2) semantic misalignment, where dense multi-scale object distributions intensify occlusion and hinder effective cross-level feature fusion. As illustrated in \cref{fig:ReduancyInVisDrone}, conventional sampling strategies introduce substantial background and similarity redundancy, overwhelming discriminative small-object cues and reducing sensitivity to fine-grained structures.

Existing efforts attempt to alleviate these challenges through architectural redesign \cite{FSAF, CEASC, RemDet}, region-level cropping \cite{SAHI, UFPMP-Det}, or resolution enhancement \cite{GLSAN, HRDNet}. However, they often neglect two critical aspects: preserving informative features under extreme redundancy and adapting feature fusion strategies to different model scales (e.g., width and depth). In UAV scenarios, valid object regions typically occupy less than 8\% of the image area \cite{VisDrone}, making redundancy suppression and spatially aligned small-object representation particularly crucial.

\begin{figure*}[t]
    \centering
    \includegraphics[width=1\textwidth]{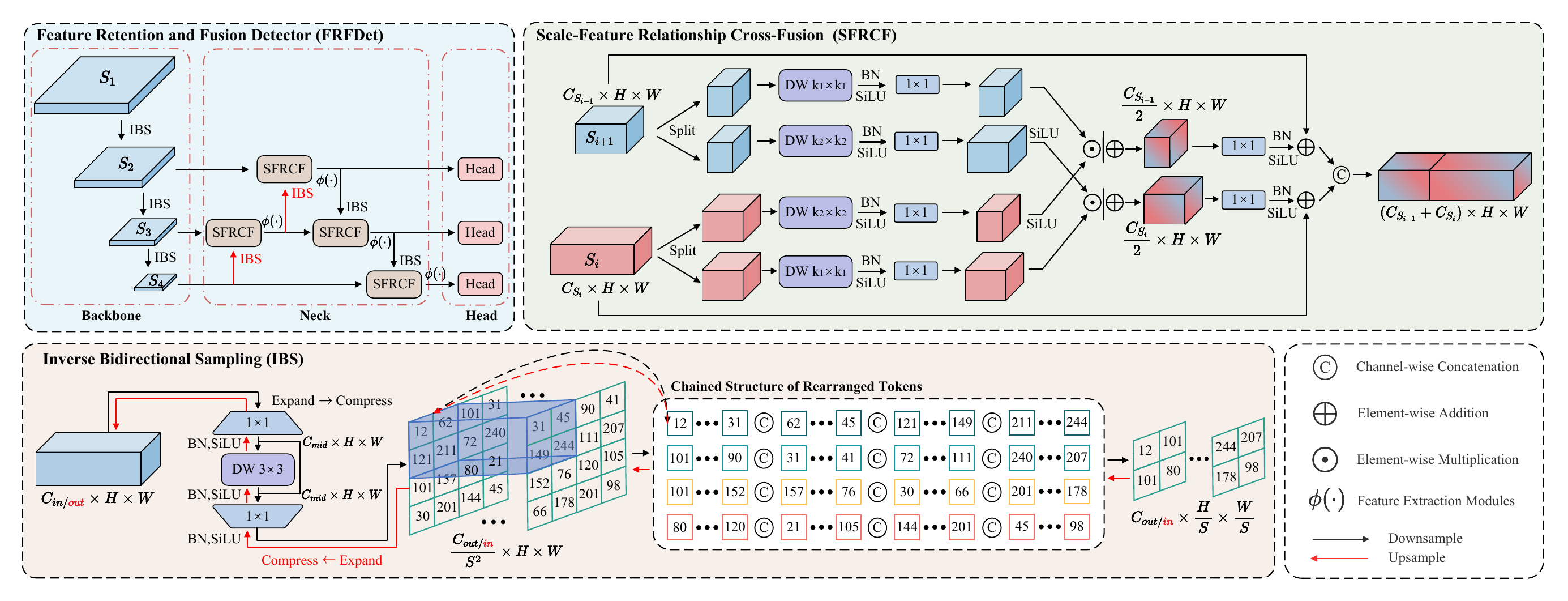}
    \caption{Overview of FRFDet. IBS enables symmetric down-/up-sampling, and SFRCF performs scale-aware cross-fusion to reduce semantic drift.}
    \label{fig:FRFDet}
\end{figure*}
To address these issues, we propose FRFDet (shown in \cref{fig:FRFDet}), a lightweight and efficient detector tailored for UAV-based small object detection. FRFDet proposes two plug-and-play modules: Inverse Bidirectional Sampling (IBS) and Scale-Feature Relationship Cross-Fusion (SFRCF). IBS adopts a symmetric expansion–compression design with channel-space reordering to suppress background redundancy while reconstructing spatial coherence, ensuring consistency between encoding and decoding stages. In contrast, SFRCF explicitly investigates the interplay between fusion strategy and model scale. We observe that inter-group element-wise multiplication enhances representation capacity in compact models by promoting nonlinear interactions among semantic groups, whereas in larger models it induces excessive entanglement due to the quadratic growth of interaction space. To this end, SFRCF adopts inter-group additive fusion for larger architectures, which preserves group-level semantics, stabilizes gradient propagation, and yields more robust fusion in cluttered UAV scenes. Extensive experiments on VisDrone \cite{VisDrone}, UAVDT \cite{UAVDT}, HazyDet \cite{HazyDet}, and MS COCO \cite{MSCOCO} show that FRFDet achieves state-of-the-art performance with lower computational cost and latency, balancing accuracy, efficiency, and scalability.

\textbf{Our main contributions are summarized as follows:}

(1) We propose FRFDet, an efficient real-time detector for UAV-based small object detection that mitigates feature redundancy and scale inconsistency.

(2) We design Inverse Bidirectional Sampling (IBS), a symmetric learnable sampling module that preserves fine-grained details and spatial alignment.

(3) We present Scale-Feature Relationship Cross-Fusion (SFRCF), a scale-aware fusion mechanism that adapts cross-level interactions to model capacity.

(4) Extensive experiments on UAV benchmarks and MS COCO demonstrate strong performance, scalability, and generalization with low computational overhead.

\section{Related Work}
\subsection{General Object Detection}
Object detectors are commonly categorized into two-stage, one-stage, and end-to-end paradigms. Two-stage methods, such as Faster R-CNN \cite{FasterRCNN}, generate region proposals followed by refinement, while one-stage detectors directly predict object locations and categories using anchor-based \cite{RetinaNet, ATSS, DDOD} or anchor-free designs \cite{Reppoints, FCOS, CenterNet, GFLV1, TOOD}. End-to-end Transformer-based detectors, including DETR \cite{DETR}, Deformable DETR \cite{DeformableDETR}, and RT-DETR \cite{RTDETR}, further simplify the pipeline by removing anchors and NMS. Recent real-time detectors, notably the YOLO series \cite{YOLOv6-3.0, YOLOv9, YOLOv10, YOLO11, YOLOv12, Gold-YOLO, HyperYOLO}, achieve strong speed–accuracy trade-offs on benchmarks such as MS COCO \cite{MSCOCO}. However, these methods often struggle in UAV scenarios, where small object size, dense distributions, and severe background redundancy degrade feature discrimination and limit generalization.

\subsection{Small Object Detection in UAV Imagery}
Small object detection in UAV imagery faces challenges from low resolution, background clutter, and limited onboard resources. Many approaches adopt coarse-to-fine strategies, using clustering or region cropping to focus on informative areas, as in ClusDet \cite{ClusDet}, DMNet \cite{DMNet}, CDMNet \cite{CDMNet}, and UFPMP-Det \cite{UFPMP-Det}. Others exploit multi-resolution features or high-resolution branches to balance local detail and global context, such as HRDNet \cite{HRDNet} and QueryDet \cite{QueryDet}, while CEASC \cite{CEASC} and ESOD \cite{ESOD} enhance detection via feature reuse or auxiliary supervision. Despite their effectiveness, these methods often introduce heuristic designs and increased architectural complexity, limiting scalability. Moreover, they largely overlook feature redundancy suppression during transformation and the impact of model scale on multi-scale fusion strategies. In contrast, our work focuses on efficient feature sampling and scale-aware fusion, providing a lightweight solution that better balances accuracy, efficiency, and scalability in UAV-based small object detection.

\section{Method}
\subsection{Rethinking Feature Generation and Fusion}\label{sec:sec3.1}
Most modern detectors adopt a multi-scale pipeline consisting of backbone feature extraction, neck aggregation, and head prediction \cite{FPN, BiFPN, YOLO11, RTDETR}. While effective for generic detection, this paradigm is suboptimal for UAV-based small object detection. In particular, small objects are highly sensitive to information loss caused by downsampling and to semantic misalignment introduced by naive cross-scale fusion. These observations motivate us to revisit feature sampling and integration from the perspective of detail preservation and scale-aware fusion.

\subsection{Inverse Bidirectional Sampling}\label{sec:sec3.2}
\begin{figure}[t]
    \centering
    \includegraphics[width=1.02\linewidth]{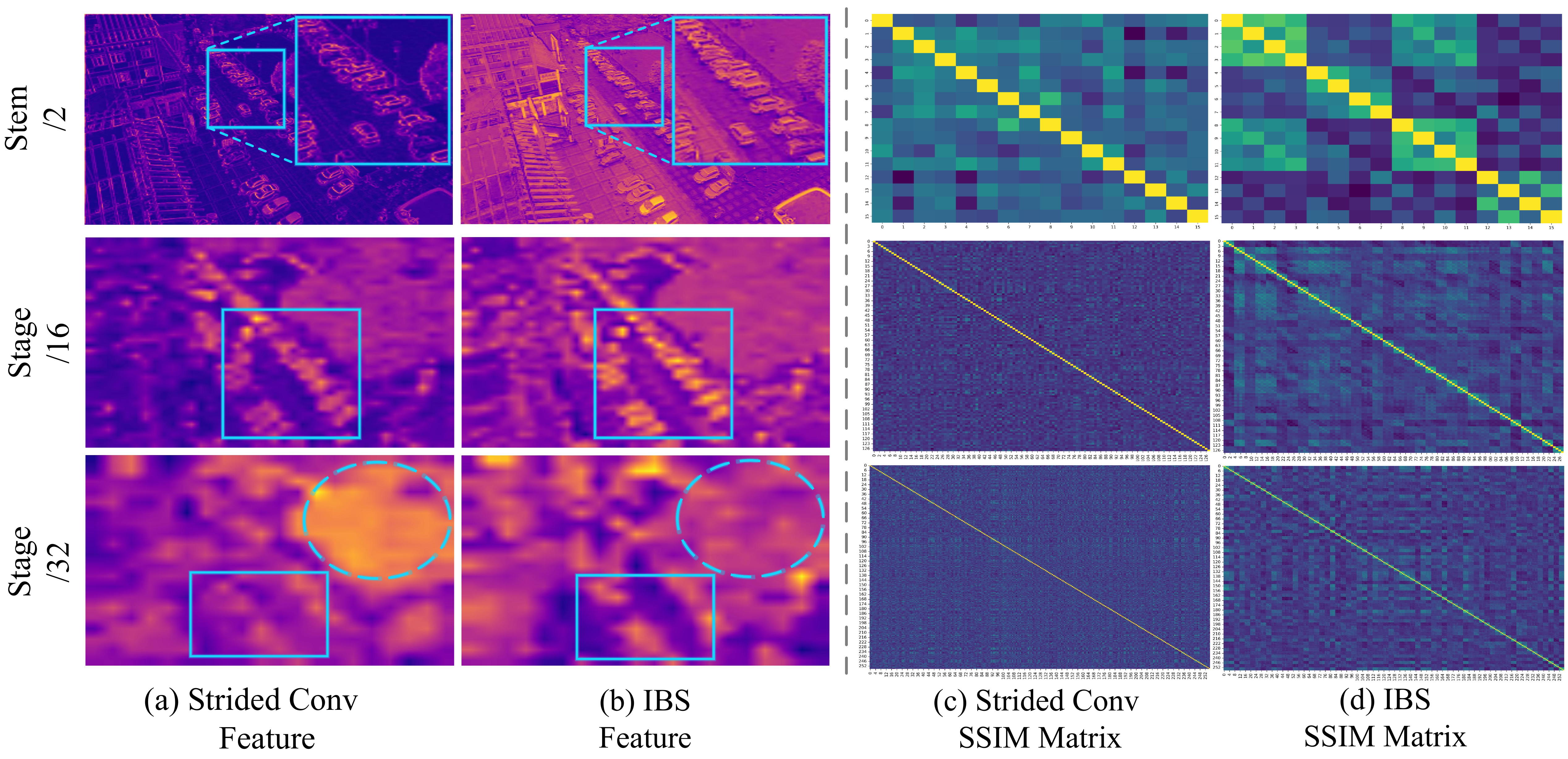}
    \caption{Feature maps and Structural Similarity Index Measure (SSIM) matrices across stages. Boxes indicate targets; brighter colors in (c) and (d) denote higher channel similarity.}
    \label{fig:IBS-Down}
\end{figure}

Conventional multi-scale feature generation relies on asymmetric operations, such as strided convolutions or pooling for downsampling \cite{FasterRCNN, TOOD, YOLOv10} and interpolation for upsampling \cite{YOLOv8, YOLO11}. Such asymmetry leads to detail degradation, background redundancy, and spatial misalignment (\cref{fig:IBS-Down} (a,c)), which are particularly harmful to small objects in UAV imagery. To address this, we propose Inverse Bidirectional Sampling (IBS), a lightweight and learnable strategy that enforces approximate structural symmetry between down- and up-sampling. Given an input feature map $X_i \in \mathbb{R}^{C_{in} \times H_i \times W_i}$, the downsampling path applies a compact expansion–compression unit with point-wise and depth-wise convolutions to enhance high-frequency details while suppressing background responses. The features are then reorganized into structured spatial groups, yielding compact and semantically coherent representations (\cref{fig:IBS-Down} (b,d)):
\begin{gather}
    \mathcal{F}_D = \mathcal{T}_{reorg} \big( \mathcal{F}_{EC}(X_i; \theta(S)) \big)
\end{gather}
where $\mathcal{F}_{EC}$ denotes the expansion–compression transformation and $\mathcal{T}_{reorg}$ performs spatial reorganization.

The upsampling process mirrors this design by applying the inverse transformation in reverse order, as indicated by the red arrows in \cref{fig:FRFDet}, reusing the same operations to reconstruct higher-resolution features:
\begin{gather}
    \mathcal{F}_U = \mathcal{F}_{EC} \big( \mathcal{T}_{reorg}(X_i; \theta(S)) \big)
\end{gather}
This symmetric formulation preserves spatial coherence across scales and mitigates semantic drift during feature transformation. Unlike interpolation-based methods or fixed rearrangement operators, IBS is fully learnable and structure-aware, enabling more stable multi-scale representations. By jointly suppressing background redundancy and preserving fine-grained details, IBS provides a robust foundation for small object detection in cluttered UAV scenes. 

\begin{figure}[t]
    \centering
    \includegraphics[width=1.01\linewidth]{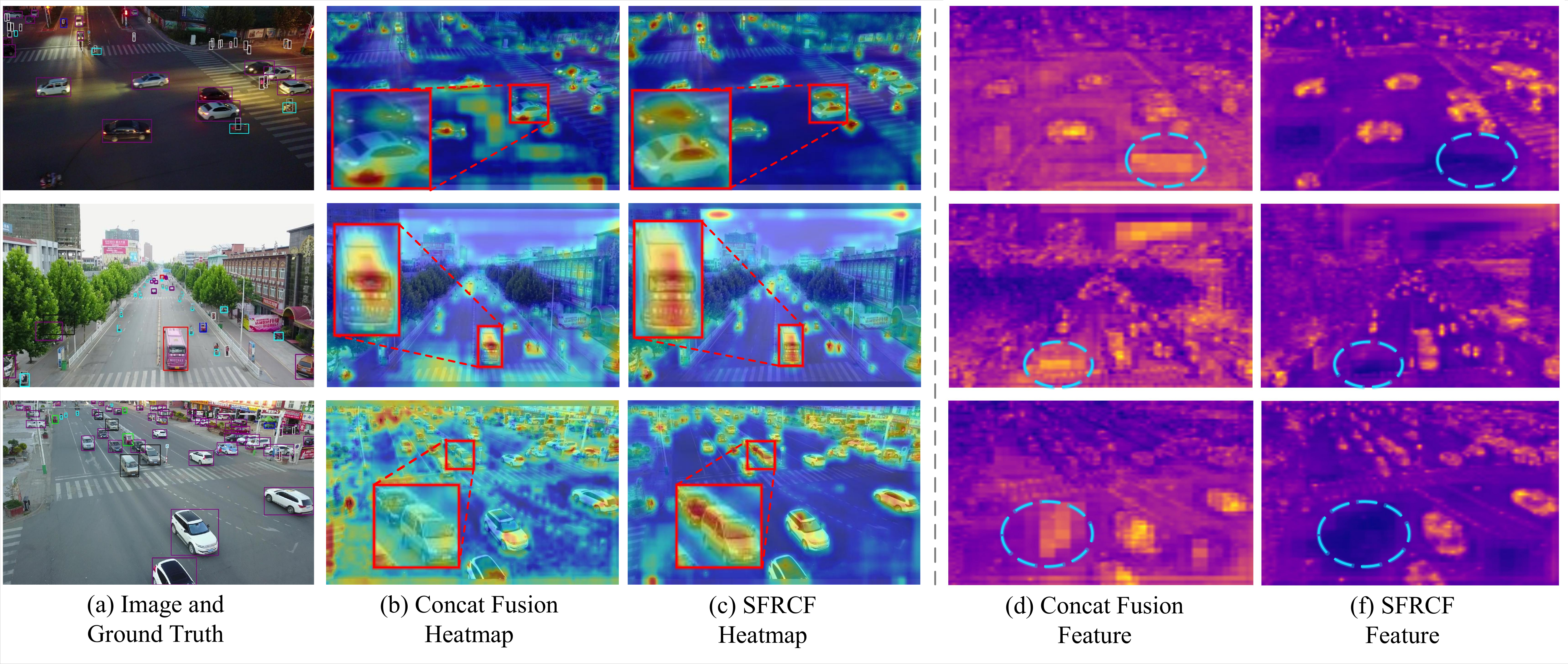}
    \caption{Fusion visualizations on VisDrone. Naive fusion shows background bias, while SFRCF enhances target features.}
    \label{fig:SFRCFVis}
\end{figure}

\begin{figure}[t]
    \centering
    \includegraphics[width=1.01\linewidth]{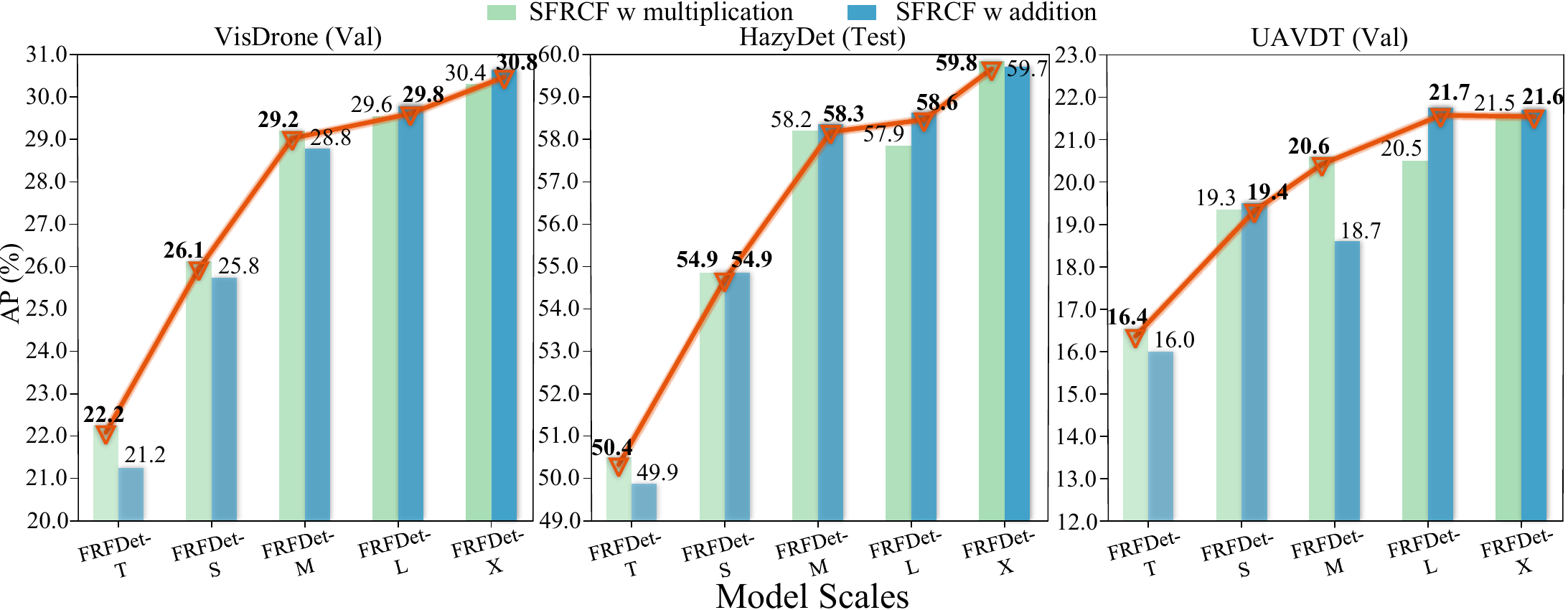}
        \caption{Comparison of AP at different scales of FRFDet on three UAV datasets, employing various fusion strategies.}
    \label{fig:smfusion-safusion}
\end{figure}

\subsection{Scale-Feature Relationship Cross-Fusion}\label{sec:sec3.3}
After refining multi-scale feature generation with IBS, we focus on cross-scale semantic fusion. Existing fusion strategies, such as concatenation, addition, and attention-based designs \cite{ViT, FPN, BiFPN, Gold-YOLO}, typically adopt fixed fusion paradigms regardless of model scale. These methods overlooks the influence of model width and depth on cross-scale interactions, which is particularly critical for small object detection in cluttered UAV scenes. Given two spatially aligned feature maps from adjacent levels, $X_{S_i} \in \mathbb{R}^{C_i \times H \times W}$ and
$X_{S_{i+1}} \in \mathbb{R}^{C_{i+1} \times H \times W}$,
SFRCF first splits them into channel groups and applies lightweight depth-wise and point-wise convolutions to enhance receptive-field diversity:
\begin{gather}
    X^{(j)}\to F^{j} = PWConv(DWConv(X^{(j)}))
\end{gather}
where $j$ denotes the group index. Let $x_1, x_2 \in \mathbb{R}^{C_{in}}$ be corresponding channel vectors from adjacent scales after transformation, and let $W_1, W_2$ denote learnable linear projections. We consider two element-wise fusion operators. Additive fusion is defined as:
\begin{gather}
    \mathcal{F}_{add} = W_{1}^{T}x_{1}+W_{2}^{T}x_{2}
\end{gather}
which aggregates features independently and scales linearly with channel dimension $C_{in}$. Multiplicative fusion is defined as:
\begin{gather}
    \mathcal{F}_{mul} = (W_{1}^{T}x_{1})\odot (W_{2}^{T}x_{2})
\end{gather}
which introduces pairwise channel interactions and yields an effective interaction space proportional to $C_{in}^2$.

This distinction reveals a scale-dependent trade-off. In compact models with limited channels, multiplicative fusion enhances expressive capacity through nonlinear cross-scale modulation, as evidenced by \cref{fig:SFRCFVis} (c) vs. (b) and \cref{fig:SFRCFVis} (f) vs. (d). However, as model width increases, the quadratic growth of interaction terms introduces redundant entanglement, amplifying background responses in cluttered UAV scenes. In contrast, additive fusion preserves group-level semantics and maintains stable feature integration in wider architectures.

Based on this analysis, SFRCF adopts a scale-aware fusion strategy, employing multiplicative fusion for compact models and additive fusion for wider models. The fused features are further refined using lightweight residual point-wise convolutions to stabilize optimization. By explicitly aligning fusion behavior with model capacity, SFRCF enables efficient and robust cross-scale semantic integration across different model scales. For more theoretical analysis, see Appendix D.

\section{Experiments}\label{sec:Experiments}

\subsection{Datasets and Metrics}\label{sec:Datasets}
We evaluate FRFDet on three UAV benchmarks—VisDrone, UAVDT, and HazyDet—as well as MS COCO 2017 to assess both task-specific effectiveness and cross-domain generalization. Following common practice \cite{CEASC, ESOD, RemDet}, results on VisDrone are reported on the validation set. UAVDT and HazyDet are evaluated on their official test splits, while COCO results are reported on the validation set under standard real-time detection protocols \cite{YOLOv10, HyperYOLO}. Performance is measured using mean Average Precision (AP), AP$_{50}$, and AP$_{75}$. Model efficiency is evaluated by parameter count, FLOPs, and FPS, measured on a single NVIDIA RTX 4090 with batch size 1.

\begin{table}[t]
    \centering
    \large
   \begin{adjustbox}{width=1.02\columnwidth,center}
        \begin{tabular}{l|c|c|cc|ccc}
        \toprule
        Method &  Input Size & Test-set & AP & AP$_{50}$ & Params (M) & FLOPs (G) & FPS\\
        \midrule
        YOLOv7-Tiny${_{\color{gray}[\text{CVPR}2023]}}$& 640$\times$640 & o & 19.1 & 34.1 & 6.0 & 6.6 & \\ 
        YOLOv8-N${_{\color{gray}[2023]}}$& 640$\times$640 & o & 19.5 & 33.3 & 3.0 & 4.3 & 233.4\\ 
        YOLOv9-T${_{\color{gray}[\text{ECCV}2024]}}$& 640$\times$640 & o & 19.7 & 33.6 & 2.0 & 3.8 & 99.4\\
        YOLOv10-N${_{\color{gray}[\text{NeurIPS}2024]}}$& 640$\times$640 & o & 19.3 & 33.1 & 2.3 & 3.3 & 158.7 \\
        YOLO11-N${_{\color{gray}[2024]}}$ & 640$\times$640& o & 19.3 & 33.1 & 2.6 & 3.2 & 177.3\\
        YOLOv12-N${_{\color{gray}[\text{arXiv}2025]}}$& 640$\times$640 & o & 19.2 & 33.2 & 2.6 & 3.2 & 143.8\\
        YOLO11-S${_{\color{gray}[2024]}}$ & 640$\times$640& o & 23.4 & 38.9 & 9.4 & 10.7 & 164.4\\
        DTSSNet${_{\color{gray}[\text{TGRS}2024]}}$& 1333$\times$800 & o & 24.2 & 39.9 & 10.1 & 50.3 & -\\
        RemDet-Tiny${_{\color{gray}[\text{AAAI}2025]}}$& 640$\times$640 & o & 21.8 & 37.1 & 3.2 & 4.6 & 289.8 \\
        RemDet-S${_{\color{gray}[\text{AAAI}2025]}}$& 640$\times$640 & o & \underline{24.7} & \underline{41.5} & 11.9 & 16.0 & -\\
        \rowcolor{tablegray}
        FRFDet-T (Ours) & 640$\times$640 & o & 22.2 & 37.2 & 2.6 & 4.6 & 144.9\\
        \rowcolor{tablegray}
        FRFDet-S (Ours) & 640$\times$640 & o & \textbf{26.1} & \textbf{42.5} & 9.3 & 15.7 & 123.7\\

        \midrule
        Faster R-CNN${_{\color{gray}[\text{TPAMI}2016]}}$& 1333$\times$800 & o & 21.4 & 40.7 & 41.8 & 187.2 & 21.1\\
        Retinanet${_{\color{gray}[\text{ICCV}2017]}}$& 1333$\times$800 & o & 21.8 & 39.3  & 38.0 & 214.7 & 19.2\\
        SAHI${_{\color{gray}[\text{ICIP}2022]}}$ & 1333$\times$800 & o & 25.6 & 43.5  & 26.1 & - & -\\
        FSAF${_{\color{gray}[\text{CVPR}2019]}}$ & 1333$\times$800 & o & 26.3 & 50.3 & - & 518.3 & 15.8\\
        YOLO11-M${_{\color{gray}[2024]}}$ & 640$\times$640 & o & 27.3 & 44.5 & 20.0 & 33.8 & 144.0\\
        RemDet-M${_{\color{gray}[\text{AAAI}2025]}}$& 640$\times$640 & o & 28.2 & 46.1 & 23.3 & 34.4 & -\\
        HRDNet${_{\color{gray}[\text{ICME}2021]}}$& 2666$\times$1600 & o & 28.3 & 49.3 & - & - & -\\
        QueryDet${_{\color{gray}[\text{CVPR}2022]}}$& 1333$\times$800 & o & 28.3 & 48.1 & - & 888.4 & 11.3 \\
        GFL V1${_{\color{gray}[\text{NeurIPS}2020]}}$& 1333$\times$800 & o & 28.4 & 50.0 & - & 525.0 & 13.5 \\
        CEASC${_{\color{gray}[\text{CVPR}2023]}}$ & 1333$\times$800 & o & 28.7 & 50.7 & - & 150.2 & 21.7 \\
        RemDet-L${_{\color{gray}[\text{AAAI}2025]}}$& 640$\times$640 & o & 29.3 & 47.4 & 35.3 & 66.7 & 138.4\\
        RemDet-X${_{\color{gray}[\text{AAAI}2025]}}$& 640$\times$640 & o & 29.9 & 48.3 & 74.1 & 112.0 & 110.2\\
        EFC${_{\color{gray}[\text{TGRS}2024]}}$& 1333$\times$800 & o & \underline{30.1} & \underline{52.1} & 40.0 & 477.6 & -\\
        \rowcolor{tablegray}
       FRFDet-M (Ours) & 640$\times$640 & o & 29.2 & 46.7 & 17.4 & 47.7 & 98.8\\
        \rowcolor{tablegray}
       FRFDet-L (Ours) & 640$\times$640 & o & 29.8 & 47.3 & 22.6 & 57.2 & 78.4\\
        \rowcolor{tablegray}
        FRFDet-X (Ours) & 640$\times$640 & o & \textbf{30.8} & 48.8 & 38.9 & 106.3 & 94.6\\
        \rowcolor{tablegray}
        FRFDet-X$^{\text{†}}$ (Ours) & 1024$\times$1024 & o  & \textbf{37.5}  & \textbf{58.4} & 38.9  & 272.2 & 45.8\\
        
        \midrule
        ClusDet${_{\color{gray}[\text{ICCV}2019]}}$ & 1000$\times$600 & o + ca & 26.7 & 50.6 & - & 436.0 & 6.5 \\
        DMNet${_{\color{gray}[\text{CVPRW}2020]}}$ & 1000$\times$600 & o + dc  & 28.2 & 47.6 & - & 471.4 & 5.9 \\
        CDMNet${_{\color{gray}[\text{ICCVW}2021]}}$ & 1000$\times$600 & ca & 29.2 & 49.5 & - & - & - \\
        GLSAN${_{\color{gray}[\text{TIP}2020]}}$ & 1333$\times$800 & o + ca & 30.7 & 55.4 & - & - & -\\ 
        AMRNet${_{\color{gray}[\text{arXiv}2020]}}$ & 1500$\times$800 & o + aug & 31.7 & 52.7 & - & - & -\\
        YOLC${_{\color{gray}[\text{TITS}2024]}}$ & 1024$\times$640 & o + ca & 31.8 & 55.0 & - & 151.0 & 2.3\\
        CZDet${_{\color{gray}[\text{CVPRW}2023]}}$ & 1999$\times$1200 & o + dc & 33.2 & 58.3 & - & - & 8.4\\
        UFPMP-Det${_{\color{gray}[\text{AAAI}2022]}}$ & 1333$\times$800 & o + ca & 36.6 & \underline{62.4} & - & 150.2 & 21.7 \\
        ESOD${_{\color{gray}[\text{TIP}2024]}}$ & 1333$\times$800 & o + ca  & 37.9 & 62.3 & - & 180.6 & 28.6 \\ 
        RemDet-X${_{\color{gray}[\text{AAAI}2025]}}$ & 1024$\times$640 & o + ca & \underline{40.0} & 61.9 & 74.1 & 182.0 & 110.2\\
       \rowcolor{tablegray}
       FRFDet-X (Ours) & 640$\times$640 & ca & \textbf{40.9} & \textbf{63.0} & 38.9 & 106.3 & 94.6\\
       \rowcolor{tablegray}
       FRFDet-X$^{\text{†}}$ (Ours) & 1024$\times$1024 & o + ca & \textbf{41.5} & \textbf{63.1} & 38.9 & 272.2 & 45.8\\

       \bottomrule
    
        \end{tabular}
    \end{adjustbox}
    \caption{Comparison with SOTA UAV detectors on VisDrone. \enquote{o}, \enquote{ca}, \enquote{dc}, and \enquote{aug} denote original, cluster-aware cropped, dense cropped, and augmented settings. \enquote{–} denotes unreported results; \enquote{†} indicates 1.6× resolution. Best results are bolded; competitors underlined; ours shaded.}
    \label{tab:VisDrone-Det}
\end{table}

\begin{table}[t]
    \centering
    \begin{adjustbox}{width=0.8\columnwidth,center}
    \begin{tabular}{l|ccc}
    \toprule
        Method & AP & AP$_{50}$ & AP$_{75}$ \\
        \midrule
        % SSD${_{\color{gray}[\text{ECCV}2016]}}$ &  9.3 & 21.4 & 6.7 \\
        Faster-RCNN${_{\color{gray}[\text{TPAMI}2016]}}$ &  11.0 & 23.4 & 8.4 \\
        CenterNet${_{\color{gray}[\text{ICCV}2019]}}$ &  13.2 & 26.7 & 11.8 \\
        ClusDet${_{\color{gray}[\text{ICCV}2019]}}$ &  13.7 & 26.5 & 12.5 \\
        DMNet${_{\color{gray}[\text{CVPRW}2020]}}$ &  14.7 & 24.6 & 12.5 \\
        % DREN${_{\color{gray}[\text{ICCVW}2019]}}$ &  15.1 & - & - \\
        CDMNet${_{\color{gray}[\text{ICCVW}2021]}}$ &  16.8 & 29.1 & 18.5 \\
        GFL V1${_{\color{gray}[\text{NeurIPS}2020]}}$ & 16.9 & 29.5 & 17.9 \\
        GLSAN${_{\color{gray}[\text{TIP}2020]}}$ &  17.0 & 28.1 & 18.8 \\
        CEASC${_{\color{gray}[\text{CVPR}2023]}}$ &  17.1 & 30.9 & 17.8 \\
        EFC${_{\color{gray}[\text{TGRS}2024]}}$ & 18.0 & 31.5 & 18.9 \\
        AMRNet${_{\color{gray}[\text{arXiv}2020]}}$ &  18.2 & 30.4 & 19.8 \\
        GLSDet${_{\color{gray}[\text{TGRS}2025]}}$ & 18.3 & 29.8 & 17.6 \\
        YOLC${_{\color{gray}[\text{TTTS}2024]}}$ &  19.3 & 30.9 & 20.1 \\
        RemDet-L${_{\color{gray}[\text{AAAI}2025]}}$ & \underline{20.6} & \underline{34.5} & \underline{22.1} \\
        \rowcolor{tablegray}
        FRFDet-L (Ours) & \textbf{21.7} & 32.7 & \textbf{25.0} \\
        \bottomrule
    \end{tabular}
    \end{adjustbox}
    \caption{SOTA comparison on UAVDT (original set).}
    \label{tab:UAVDT}
\end{table}

\begin{table}[t]
    \centering
    \large
    \begin{adjustbox}{width=0.99\columnwidth,center}
    \begin{tabular}{l|cc|cc|cc}
    \toprule
        \multirow{2}{*}{Method} & \multicolumn{2}{c|}{Test-set} & \multicolumn{2}{c|}{RDDTS} & \multirow{2}{*}{Params (M)} & \multirow{2}{*}{FLOPs (G)}\\
        \cmidrule{2-5}
        & AP & AP$_{50}$ & AP & AP$_{50}$ & & \\
        
        \midrule
        \midrule
        \multicolumn{7}{l}{\textit{Mode: Pretrained on ImageNet-1K + fine-tuned with the “1$\times$” schedule (12 epochs)}} \\
        \midrule

         Dynamic R-CNN${_{\color{gray}[\text{ECCV}2020]}}$ & 47.6 & 63.5 & 22.5 & 32.4 & 41.4 & 201.7 \\
        Grid R-CNN${_{\color{gray}[\text{CVPR}2019]}}$ & 50.5 & 68.6 & 25.2 & 36.6 & 64.5 & 317.4 \\

        \midrule
        GFL V1${_{\color{gray}[\text{NeurIPS}2020]}}$ & 36.8 & 52.2 & 13.9 & 22.9 & 32.3 & 198.7\\
        Reppoints${_{\color{gray}[\text{ICCV}2019]}}$ & 43.8 & 65.7 & 21.3 & 35.1 & 36.8 & 184.3 \\
        FCOS${_{\color{gray}[\text{ICCV}2019]}}$ & 45.9 & 67.1 & 22.8 & 34.5 & 32.1 & 191.5\\
        CenterNet${_{\color{gray}[\text{ICCV}2019]}}$ & 47.2 & 67.0 & 23.8 & 37.1 & 32.1 & 191.5\\
        ATSS${_{\color{gray}[\text{CVPR}2020]}}$ & 50.4 & 68.4 & 25.1 & 37.7 & 32.1 & 195.6\\
        DDOD${_{\color{gray}[\text{ACM MM}2021]}}$ & 50.7 & 68.6 & 26.1 & 38.6 & 32.2 & 173.1\\
        TOOD${_{\color{gray}[\text{CVPR}2021]}}$ & 51.4 & 69.8 & 25.8 & 39.0 & 32.0 & 192.5\\
        DeCoDet${_{\color{gray}[\text{arXiv}2024]}}$ & 51.5 & 71.1 & 25.9 & 39.5 & 34.6 & 225.4 \\
        
        \midrule
        Deformable DETR${_{\color{gray}[\text{ICLR}2021]}}$ & 51.9 & 71.5 & 26.5 & 40.2 & 40.0 & 192.5 \\
        \rowcolor{tablegray}
        \midrule
        FRFDet-P (Ours) & \textbf{54.4} & \textbf{75.1} & \textbf{27.2} & \textbf{41.3} & 33.9 & 228.2 \\
        
        \midrule
        \midrule
        \multicolumn{7}{l}{\textit{Mode: Trained from scratch for 300 epochs}} \\
        \midrule

        YOLOv8-N${_{\color{gray}[2023]}}$ & 47.5 & 67.8 & 25.5 & 39.1 & 3.0 & 4.0 \\
        YOLOv9-T${_{\color{gray}[\text{ECCV}2024]}}$ & 47.6 & 67.7 & 26.8 & 40.6 & 2.0 & 3.8\\
        YOLOv10-N${_{\color{gray}[\text{NeurIPS}2024]}}$ & 47.5 & 67.6 & 26.7 & 41.2 & 2.3 & 3.3\\
        YOLO11-N${_{\color{gray}[2024]}}$ & 47.5 & 67.7 & 26.7 & 40.9 & 2.6 & 3.2\\
        YOLOv12-N${_{\color{gray}[\text{arXiv}2024]}}$ & 47.5 & 68.0 & 25.7 & 39.8 & 2.6 & 3.2 \\
        
        \rowcolor{tablegray}
        FRFDet-T (Ours) & 50.4 & 69.9 & 28.5 & 42.6 & 2.6 & 4.6 \\
        \rowcolor{tablegray}
        FRFDet-S (Ours) & 54.9 & 74.4 & 32.4 & 47.2 & 9.3 & 15.7 \\
        \rowcolor{tablegray}
        FRFDet-M (Ours) & 58.3 & 77.8 & 36.6 & 52.1 & 17.4 & 47.7 \\
        \rowcolor{tablegray}
        FRFDet-L (Ours) & 58.6 & 77.9 & 37.8 & 53.7 & 22.7 & 57.2 \\
        \rowcolor{tablegray}
        FRFDet-X (Ours) & \textbf{59.8} & \textbf{79.0} & \textbf{38.2} & \textbf{54.4} & 39.0 & 106.3 \\
        
        \bottomrule
    \end{tabular}
    \end{adjustbox}
    \caption{Results on HazyDet and RDDTS. ResNet-50 pretrained; FRFDet-P built on TOOD.}
    \label{tab:HazyDet}
\end{table}

\begin{table}[h]
    \centering
    \large
    \begin{adjustbox}{width=0.99\columnwidth,center}
    \begin{tabular}{l|cc|cc}
    \toprule
        Method & AP & AP$_{50}$ & Params (M) & FLOPs (G) \\
    \midrule
    YOLOv6-3.0-N${_{\color{gray}[\text{arXiv}2023]}}$ & 37.5 & 52.7 & 4.7 & 5.7 \\
    YOLOv7-Tiny${_{\color{gray}[\text{CVPR}2023]}}$ & 37.4 & 55.2 & 6.2 & 6.8 \\
    YOLOv8-N${_{\color{gray}[2023]}}$ & 37.3 & 52.6 & 3.2 & 4.4 \\
    Gold-YOLO-N${_{\color{gray}[\text{NeurIPS}2023]}}$ & 39.6 & 55.7 & 5.6 & 6.0 \\
    YOLOv9-T${_{\color{gray}[\text{ECCV}2024]}}$ & 38.3 & 53.1 & 2.0 & 3.8 \\
    YOLOv10-N${_{\color{gray}[\text{NeurIPS}2024]}}$ & 38.5 & 52.8 & 2.3 & 3.3\\
    Hyper-YOLO-T${_{\color{gray}[\text{TPAMI}2024]}}$ & 38.5 & 54.5 & 3.1 & 4.8\\
    YOLO11-N${_{\color{gray}[2024]}}$ & 39.5 & 55.1 & 2.6 & 3.2\\
    RemDet-Tiny${_{\color{gray}[\text{AAAI}2025]}}$ & 39.5 & 55.8 & 3.2 & 4.6\\
    YOLOv12-N${_{\color{gray}[\text{arXiv}2025]}}$ & \underline{40.6} & \underline{56.7} & 2.6 & 3.2\\
    \rowcolor{tablegray}
    FRFDet-T (Ours) & \textbf{41.3} & \textbf{57.2} &  2.6 & 4.6\\

    \midrule
    YOLOv6-3.0-S${_{\color{gray}[\text{arXiv}2023]}}$ & 44.3 & 61.2 & 18.5 & 22.6\\
    YOLOv8-S${_{\color{gray}[2023]}}$ & 44.9 & 61.8 & 11.2 & 14.3 \\ 
    Gold-YOLO-S${_{\color{gray}[\text{NeurIPS}2023]}}$ & 45.4 & 62.5 & 21.5 & 23.0\\
    RemDet-S${_{\color{gray}[\text{AAAI}2025]}}$ & 45.5 & 62.8 & 11.9 & 16.0 \\
    YOLOv10-S${_{\color{gray}[\text{NeurIPS}2024]}}$ & 46.3 & 63.0 & 7.2 & 10.8 \\
    RT-DETR-R18${_{\color{gray}[\text{CVPR}2024]}}$ & 46.5 & 63.8 & 20.2 & 60.0 \\
    YOLOv9-S${_{\color{gray}[\text{ECCV}2024]}}$ & 46.8 & 63.4 & 7.1 & 13.2 \\
    YOLO11-S${_{\color{gray}[2024]}}$ & 47.0 & 63.8 & 9.4 & 10.8 \\
    Hyper-YOLO-S${_{\color{gray}[\text{TPAMI}2024]}}$ & \underline{48.0} & \underline{65.1} & 14.8 & 19.5\\
    YOLOv12-S${_{\color{gray}[\text{arXiv}2025]}}$ & \underline{48.0} & 65.0 & 9.3 & 10.7\\
    \rowcolor{tablegray}
    FRFDet-S (Ours) & \textbf{48.3} & \textbf{65.4} & 9.3 & 15.8\\
    
    \bottomrule
    \end{tabular}
    \end{adjustbox}
    \caption{Comparison with real-time detectors on MS COCO.}
    \label{tab:MSCOCO}
\end{table}

\begin{table}[t]
    \centering
    \large
    \begin{adjustbox}{width=0.99\columnwidth,center}
    \begin{tabular}{ccc|cc|ccc}
    \toprule
        IBS-D & IBS-U & SFRCF & AP & AP$_{50}$ & Params & FLOPs\\
    \midrule
        - & - & - & 19.3 \footnotesize\textcolor{gray}{+(0.0)} & 33.1 \footnotesize\textcolor{gray}{+(0.0)} & 2.6 & 3.2 \\
    \midrule
        \checkmark & - & - & 20.7 \footnotesize\textcolor{tableup}{+(1.4)} & 35.0 \footnotesize\textcolor{tableup}{+(1.9)} & 2.1 & 3.6 \\
        - & \checkmark & - & 20.0 \footnotesize\textcolor{tableup}{+(0.7)} & 34.1 \footnotesize\textcolor{tableup}{+(1.0)} & 2.8 & 3.7 \\
        \checkmark & \checkmark & - & 21.5 \footnotesize\textcolor{tableup}{+(2.2)} & 35.9 \footnotesize\textcolor{tableup}{+(2.8)} & 2.3 & 4.1\\
    \midrule
        - & - & \checkmark & 20.2 \footnotesize\textcolor{tableup}{+(0.9)} & 34.3 \footnotesize\textcolor{tableup}{+(1.2)} & 2.8 & 3.6 \\
        \checkmark & \checkmark & \checkmark & 22.2 \footnotesize\textcolor{tableup}{+(2.9)} & 37.2 \footnotesize\textcolor{tableup}{+(4.1)} & 2.6 & 4.6\\
    
    \bottomrule
    \end{tabular}
    \end{adjustbox}
    \caption{Component ablation of FRFDet-T on VisDrone.}
    \label{tab:ablation_all}
\end{table}

\subsection{Implementation Details}\label{sec:Implementation}
All models are implemented with Ultralytics and MMDetection. FRFDet is trained from scratch using YOLO11. For UAV benchmarks, models are trained for 300 epochs with SGD (lr = 0.01, batch = 16). For MS COCO, the batch size is set to 32, with other settings unchanged. All experiments use a 640×640 input unless specified.

\subsection{Comparative Results and Analysis}\label{sec:Main Results}
\subsubsection{VisDrone} As shown in \cref{tab:VisDrone-Det}, FRFDet consistently outperforms state-of-the-art lightweight detectors. FRFDet-S surpasses YOLO11-S, DTSSNet \cite{DTSSNet}, and RemDet-S \cite{RemDet} by 1.4–2.7\% AP with fewer parameters and lower FLOPs. At larger scales, FRFDet-X achieves 30.8\% AP with 94.6 FPS, outperforming RemDet-X while using nearly half the parameters. Performance further improves with higher resolution, demonstrating the robustness of IBS across scales. Under cluster-based cropping, FRFDet achieves up to 41.5\% AP, confirming its strong compatibility without relying on handcrafted region filtering.
\subsubsection{UAVDT}
On UAVDT (\cref{tab:UAVDT}), FRFDet-L achieves 21.7\% AP, outperforming previous UAV-specific detectors such as YOLC \cite{YOLC}, EFC \cite{EFC}, GLSDet \cite{GLSDet}, and RemDet-L. This highlights the effectiveness of scale-aware sampling and fusion in densely packed traffic scenes.
\subsubsection{HazyDet}
FRFDet shows strong robustness under degraded visual conditions (\cref{tab:HazyDet}). In pretraining–finetuning settings, FRFDet improves over TOOD \cite{TOOD} by +3.0\% AP on synthetic data and +1.4\% on real foggy scenes. When trained from scratch, FRFDet consistently outperforms YOLOv8–YOLOv12, and FRFDet-X establishes a new state-of-the-art with 59.8\% AP, demonstrating resilience to severe visibility degradation.
\subsubsection{MS COCO}
On MS COCO (\cref{tab:MSCOCO}), FRFDet generalizes well beyond UAV scenarios. In particular, compact variants (T/S) outperform recent real-time detectors by up to +1.8\% AP, validating the transferability of IBS and SFRCF to generic object detection.

\subsection{Ablation and Analysis}\label{sec:Main Results}
\subsubsection{Effect of IBS and SFRCF}
Ablations on VisDrone (\cref{tab:ablation_all}) show that symmetric IBS improves AP by +2.2\%, confirming its effectiveness in preserving spatial structure and suppressing redundancy. Adding SFRCF brings a further +0.9\% AP gain on compact models, validating scale-aware fusion. Additional analyses are in Appendices C and D.

\subsubsection{Fusion Strategy vs. Model Scale}
As illustrated in \cref{fig:smfusion-safusion}, element-wise multiplication consistently outperforms addition in compact models, while additive fusion is more effective in wider architectures. This trend is more pronounced when increasing model width, confirming that fusion behavior is strongly coupled with model capacity, as assumed in SFRCF.

\subsubsection{Visualization}
Qualitative results on VisDrone, HazyDet, and MS COCO show that FRFDet detects more objects with less category confusion than prior general and small-object SOTA methods (see Appendix E).

\section{Conclusion}
This paper presents FRFDet, an efficient single-stage detector for UAV-based small object detection. To mitigate feature redundancy and semantic inconsistency in multi-scale representations, we develop IBS, a symmetric sampling strategy for preserving fine-grained details, and SFRCF, a scale-aware cross-group fusion mechanism that adapts fusion behaviors to model capacity. Extensive experiments on UAV and generic benchmarks demonstrate strong effectiveness, scalability, and generalization, indicating that FRFDet provides a practical and deployable solution for real-time aerial perception.

\bibliographystyle{IEEEtran}
\bibliography{icme2026references}

\clearpage
% \appendix

\setcounter{page}{1} % 
\renewcommand{\thesection}{} % 取消附录部分的章节编号 
\renewcommand{\thesection}{\Alph{section}} 
\setcounter{section}{0} 
\renewcommand{\thetable}{\Roman{table}}
\setcounter{table}{5} % 从 1 开始编号附录中的表格 
\setcounter{figure}{5} % 从 1 开始编号附录中的图像
\setcounter{equation}{5}

% \setcounter{section}{0}
% \renewcommand{\thesection}{\Alph{section}}

% \setcounter{table}{0}
% \setcounter{figure}{0}
% \setcounter{equation}{0}

% \renewcommand{\thetable}{\Alph{section}.\arabic{table}}
% \renewcommand{\thefigure}{\Alph{section}.\arabic{figure}}
% \renewcommand{\theequation}{\Alph{section}.\arabic{equation}}

% \documentclass[conference]{IEEEtran}
% \IEEEoverridecommandlockouts
% The preceding line is only needed to identify funding in the first footnote. If that is unneeded, please comment it out.
% \usepackage{cite}
% \usepackage{amsmath,amssymb,amsfonts}
% \usepackage{algorithmic}
% \usepackage{graphicx}
% \usepackage{textcomp}
% \usepackage{xcolor}
% \usepackage{hyperref}
% \usepackage{cleveref}
% \usepackage{colortbl}
% \usepackage{xcolor}
% \usepackage{booktabs}
% \usepackage{adjustbox}
% \usepackage{caption}
% \usepackage{csquotes}
% \usepackage{multirow}
\definecolor{tablegray}{HTML}{D8D8D8}
\definecolor{tableup}{RGB}{191,121,64}
\definecolor{tabledown}{HTML}{D8D8D8}
\def\BibTeX{{\rm B\kern-.05em{\sc i\kern-.025em b}\kern-.08em
    T\kern-.1667em\lower.7ex\hbox{E}\kern-.125emX}}

% \title{Appendix}

% \maketitle

\section{Architectural Scaling Analysis: Width and Depth Factors}
\label{sec:reference_examples}
We construct FRFDet models at different scales by adjusting the width and depth factors, as summarized in \cref{tab:model_factor}, to validate the scalability of our architecture. To examine how model scale influences the choice of fusion strategy, we conduct two sets of controlled experiments: (1) For FRFDet-T/S/M/X, we fix the depth factor at 0.5 and progressively increase the width factor from 0.25 to 1.5; (2) For FRFDet-M and FRFDet-L, we fix the width factor at 1.0 and gradually increase the depth factor from 0.5 to 1.0.

Based on the results presented in fig. 5 and section IV.D of the main text, we observe a clear correlation between the preferred fusion strategy and the model capacity. Specifically, inter-group element-wise multiplication facilitates non-linear pairwise interactions, which compensates for the limited representation capacity of lightweight models (FRFDet-T/S), particularly under challenging conditions such as low-light or foggy UAV imagery. In contrast, inter-group element-wise addition benefits larger models (FRFDet-L/X) by leveraging identity mapping to mitigate background redundancy caused by channel over-expansion. Interestingly, despite the adoption of different fusion strategies across scales, increasing the width factor consistently yields greater performance gains than increasing the depth factor, which aligns with the scale law phenomenon \cite{ScaleLaw}. 

\begin{table}[h]
    \centering
    \begin{adjustbox}{width=0.72\columnwidth,center}
    \begin{tabular}{l|cc}
    \toprule
    Model Scale & Width Factor & Depth Factor\\
     \midrule 
        FRFDet-T & 0.25 & 0.5  \\
        FRFDet-S & 0.5 & 0.5 \\
        FRFDet-M & 1.0 & 0.5 \\
        FRFDet-L & 1.0 & 1.0 \\
        FRFDet-X & 1.5 & 0.5 \\
    \bottomrule
         
    \end{tabular}
    \end{adjustbox}
    \caption{Settings for the ratio of model scale factors.}
    \label{tab:model_factor}
\end{table}

Beyond empirical observations, we further provide a theoretical perspective in Appendix D, analyzing the entropy of foreground and background features to better explain the effectiveness of each fusion strategy.

\section{Details of IBS}
This section details the design of the Inverse Bidirectional Sampling (IBS) module, which addresses the critical challenge of excessive background redundancy in UAV imagery that weakens small object features. IBS integrates a channel expansion-compression unit $\mathcal{F}_{EC}$ and a spatial reorganization operator $\mathcal{T}_{reorg}$ to retain crucial small object details while reducing redundancy. Its bidirectional and approximately symmetric design further enhances local spatial consistency and representation fidelity. To provide a comprehensive understanding, this section extends the formulations of Eq. (1) and Eq. (2) in the main text.

Let the input feature map be $X \in \mathbb{R}^{C_{in} \times H \times W}$. The $\mathcal{F}_{EC}$ is designed to enhance cross-channel interaction and suppress redundancy by employing 1×1 convolutions for dimensional expansion and compression. To enrich the representation capacity in the high-dimensional space, small-kernel depthwise convolutions are inserted to capture localized high-frequency information. The formulation of the downsampling path IBS-D is as follows:
\begin{gather}
    \mathcal{F}_{DW}^{E} = \text{SiLU}\big(\text{BN}(\text{Conv}_{1\times1}^{C_{in} \rightarrow{C_{mid}}}(X))\big) \\
    \mathcal{F}_{PW}^{IBS} = \text{SiLU}\big(\text{BN}(\text{DWConv}_{3\times3}^{C_{mid}\rightarrow{C_{mid}}}(X))\big) \\
    \label{eq:ibs_down}
    \mathcal{F}_{DW}^{C} = \text{SiLU}\big(\text{BN}(\text{Conv}_{1\times1}^{C_{mid} \rightarrow{C_{\frac{out}{S^{2}}}}}(X))\big) \\
    X^{'} = \mathcal{F}_{DW}^{E}(X) \\
    \mathcal{F}_{EC} = Y = \mathcal{F}_{DW}^{C}(\mathcal{F}_{PW}^{IBS}(X^{'}) + X^{'})
\end{gather}

\begin{table}[t]
    \centering
    \begin{adjustbox}{width=0.8\columnwidth,center}
    \begin{tabular}{l|cc|ccc}
    \toprule
        $r$ &  AP & AP$_{50}$ & Params (M) & FLOPs (G) & FPS\\
        \midrule
        1 & 21.4 & 36.5 & \textbf{2.5} & \textbf{3.9} & \textbf{148.2}\\
        2 & \textbf{22.2} & \textbf{37.2} & 2.6 & 4.6 & 144.9 \\
        4 & \textbf{22.2} & 37.0 & 2.8 & 5.9 & 115.3 \\
        \bottomrule
    \end{tabular}
    \end{adjustbox}
    \caption{Ablation study on expansion ratios $r$ in IBS for FRFDet-T on VisDrone.}
    \label{tab:IBS-R}
\end{table}

\begin{table}[t]
    \centering
    \begin{adjustbox}{width=0.86\columnwidth,center}
    \begin{tabular}{l|cc}
    \toprule
        Method & AP & AP$_{50}$\\
        \midrule
        IBS-D w/o Residual + IBS-U w/o Residual & 21.7 & 36.5\\
        IBS-D w/o Residual + IBS-U w Residual & 21.8 & 36.7 \\
        IBS-D w Residual + IBS-U w/o Residual & \textbf{22.2} & \textbf{37.2} \\
        IBS-D w Residual + IBS-U w Residual & 22.0 & 37.1 \\
        \bottomrule
    \end{tabular}
    \end{adjustbox}
    \caption{Ablation study on residual connections in IBS for FRFDet-T on VisDrone.}
    \label{tab:IBS-Residual}
\end{table}

\noindent
Here, $\mathcal{F}_{DW}^{E}$ and $\mathcal{F}_{DW}^{C}$ denote the expansion and compression operations, respectively. The intermediate channel dimension is set to $C_{mid} = C_{out} \times r$, where $r$ is the expansion ratio. As shown in the ablation study in \cref{tab:IBS-R}, $r=2$ achieves the best trade-off between detection accuracy and computational cost.

Following $\mathcal{F}_{EC}$, a spatial reorganization step is applied. The sampling factor $S$ (typically 2) is used to compress the channels to $\frac{C{out}}{S^2}$ and prepare the feature map for spatial rearrangement. Specifically, the tensor $Y \in \mathbb{R}^{\frac{C_{out}}{S^2} \times H \times W}$ is first unfolded with a sliding window of stride $S$:
\begin{gather}
Y{'} = \mathcal{F}_{unfold}(Y)
\end{gather}

\noindent
Then, the $S \times S$ local patches are flattened and merged into the channel dimension, yielding:
\begin{gather}
\mathcal{T}_{reorg} = Y^{''} = \mathcal{F}_{rearrange}(Y^{'}) \in \mathbb{R}^{C_{out} \times H' \times W'}
\end{gather}

\noindent
This completes the IBS-D downsampling path. Motivated by the observed drawbacks of asymmetric sampling in feature reconstruction and inter-channel SSIM (see fig. 3(a,c)), we propose a structurally symmetric upsampling variant (IBS-U). It follows the IBS-D design while discarding the residual connection in $\mathcal{F}_{EC}$:
\begin{gather}
Y{'} = \mathcal{F}_{rearrange}(Y^{''}) \\
Y = \mathcal{F}_{fold}(Y^{'}) \\
X = \mathcal{F}_{DW}^{C}(\mathcal{F}_{PW}^{IBS}(\mathcal{F}_{DW}^{E}(Y)))
\end{gather}

\noindent
A key difference between IBS-U and IBS-D lies in the use of residual connections. In IBS-U, residual connections are preserved to facilitate the propagation of low-level detail features from earlier layers. For high-resolution UAV imagery, retaining fine-grained details is critical for accurate small object detection and classification. By injecting uncompressed shallow features directly into deeper layers, the residual path helps sustain fine-grained information flow throughout the network. In contrast, IBS-D typically operates in the deeper decoder layers. Introducing abstract semantic features through residual connections at this stage may disrupt spatial consistency when fusing with shallow features, potentially leading to localization bias. To validate this design choice, we conduct an ablation study on different combinations of residual connections in IBS-D and IBS-U. As shown in \cref{tab:IBS-Residual}, keeping the residual connection in IBS-D while removing it from IBS-U yields the best performance, further validating the soundness of our analysis.

\section{Further Comparison and Ablation Studies of FRFDet}

IBS-D/U outperforms conventional sampling methods (e.g., PatchMerging \cite{Swin}, SoftPool \cite{SoftPool}, ConditionalPool \cite{ConditionalPool}, CARAFE \cite{CARAFE}, Dysample \cite{Dysample}), as shown in \cref{tab:ablation_IBS}, due to the joint effect of enhanced detail preservation and background suppression.

As shown in \cref{tab:ablation_SFRCF}, SFRCF surpasses AFPN \cite{AFPN} and CFP \cite{CFP} with higher accuracy and lower FLOPs. Unlike direct addition (AFPN) or top-down semantic injection (CFP), SFRCF utilizes cross-group element-wise multiplication for nonlinear interactions, enhancing representation in compact models.

To explore the impact of kernel size design in cross-scale fusion (SFRCF), we investigate whether globally fixed kernel sizes or hierarchically adaptive kernel sizes are more effective in capturing multi-scale semantics. This is motivated by the inherent disparity between shallow and deep features—where shallow layers encode fine-grained spatial details and deeper layers capture abstract global semantics.

For fixed-size kernels, we evaluate standard $3\times3$ and $5\times5$ configurations across all SFRCF stages. For hierarchically adaptive kernels, we fix one side of the convolution kernel at $3\times3$ while varying the other side $k_2 \times k_2$ according to layer depth:
\begin{gather}
k_2 = 3 \times (l_0 + \Delta l \cdot (l - l_0))
\end{gather}

\noindent
where $l_0$ denotes the starting layer, $\Delta l$ is the kernel growth factor, and $l$ is the current layer index. This adaptive strategy aligns the receptive field with the semantic abstraction level at each stage.

As shown in \cref{tab:SFRCF_k}, the hierarchically adaptive kernel configuration outperforms both fixed-size baselines, achieving the highest AP and AP$_{50}$ with negligible increases in parameters and FLOPs. These results highlight the advantage of depth-aware receptive field design in fusing neighboring-scale features and demonstrate its effectiveness in enhancing detection performance.

\begin{table}[t]
    \centering
    \large
    \begin{adjustbox}{width=0.99\columnwidth,center}
    \begin{tabular}{l|l|cc|cc}
    \toprule
        Sampling & Method & AP & AP$_{50}$ & Params & FLOPs\\
    \midrule
        \multirow{4}{*}{Downsampling} & PatchMerging${_{\color{gray}[\text{ICCV}2021]}}$ & 18.3 & 31.7 & 2.2 & 2.7 \\
        & SoftPool${_{\color{gray}[\text{ICCV}2021]}}$ & 19.2 & 32.8 & 1.9 & 2.4 \\
        & ConditionalPool${ _{\color{gray}[\text{PR}2023]}}$ & 19.2 & 33.0 & 1.9 & 2.4  \\
        & \cellcolor{tablegray}IBS-D (Ours) & \cellcolor{tablegray}\textbf{20.7} & \cellcolor{tablegray}\textbf{35.0} & \cellcolor{tablegray}2.1 & \cellcolor{tablegray}3.6 \\
        \midrule
        \multirow{4}{*}{Upsampling} & Nearest & 19.3 & 33.1 & 2.6 & 3.2\\
        & CARAFE${_{\color{gray}[\text{ICCV}2019]}}$ &  19.4 & 33.0 & 2.9 & 3.6\\
        & Dysample${_{\color{gray}[\text{ICCV}2023]}}$ & 19.7 & 33.7 & 2.6 & 3.2 \\
        & \cellcolor{tablegray}IBS-U (Ours) & \cellcolor{tablegray}\textbf{20.0} & \cellcolor{tablegray}\textbf{34.1} & \cellcolor{tablegray}2.8 & \cellcolor{tablegray}3.7 \\
    \bottomrule
    \end{tabular}
    \end{adjustbox}
    \caption{Comparison of upsampling and downsampling methods in IBS with other approaches on VisDrone.}
    \label{tab:ablation_IBS}
\end{table}

\begin{table}[t]
    \centering
    \large
    \begin{adjustbox}{width=0.99\columnwidth,center}
    \begin{tabular}{l|l|cc|cc}
    \toprule
        Fusion & Method & AP & AP$_{50}$ & Params & FLOPs\\
    \midrule
        Direct fusion & AFPN${_{\color{gray}[\text{SMC}2023]}}$ & 19.1 & 32.6 & 2.5 & 4.1 \\
        \midrule
        \multirow{2}{*}{Progressive fusion} & CFP${_{\color{gray}[\text{TIP}2023]}}$ & 19.5 & 33.6 & 6.9 & 9.8 \\
         & \cellcolor{tablegray}SFRCF (Ours) & \cellcolor{tablegray}\textbf{20.2} & \cellcolor{tablegray}\textbf{34.3} & \cellcolor{tablegray}2.8 & \cellcolor{tablegray}3.6 \\
    \bottomrule
    \end{tabular}
    \end{adjustbox}
    \caption{Performance comparison of SFRCF with other direct fusion method and progressive fusion method on VisDrone.}
    \label{tab:ablation_SFRCF}
\end{table}

\begin{table}[t]
    \centering
    \begin{adjustbox}{width=1.0\columnwidth,center}
    \begin{tabular}{l|cc|cc}
    \toprule
        Kernel Sizes &  AP & AP$_{50}$ & Params (M) & FLOPs (G)\\
        \midrule
        (3,3), (3,3), (3,3), (3,3) & 21.7 & 36.5 & 2.6 & 4.6 \\
        (5,5), (5,5), (5,5), (5,5) & 21.8 & 36.8 & 2.6 & 4.6\\
        (3,5), (3,3), (3,5), (3,7) & \textbf{22.2} & \textbf{37.2} & 2.6 & 4.6 \\
        \bottomrule
    \end{tabular}
    \end{adjustbox}
    \caption{Ablation study on kernel sizes for neighboring scale fusion across SFRCF stages in FRFDet-T on VisDrone.}
    \label{tab:SFRCF_k}
\end{table}

\begin{figure*}
    \centering
    \includegraphics[width=1.0\linewidth]{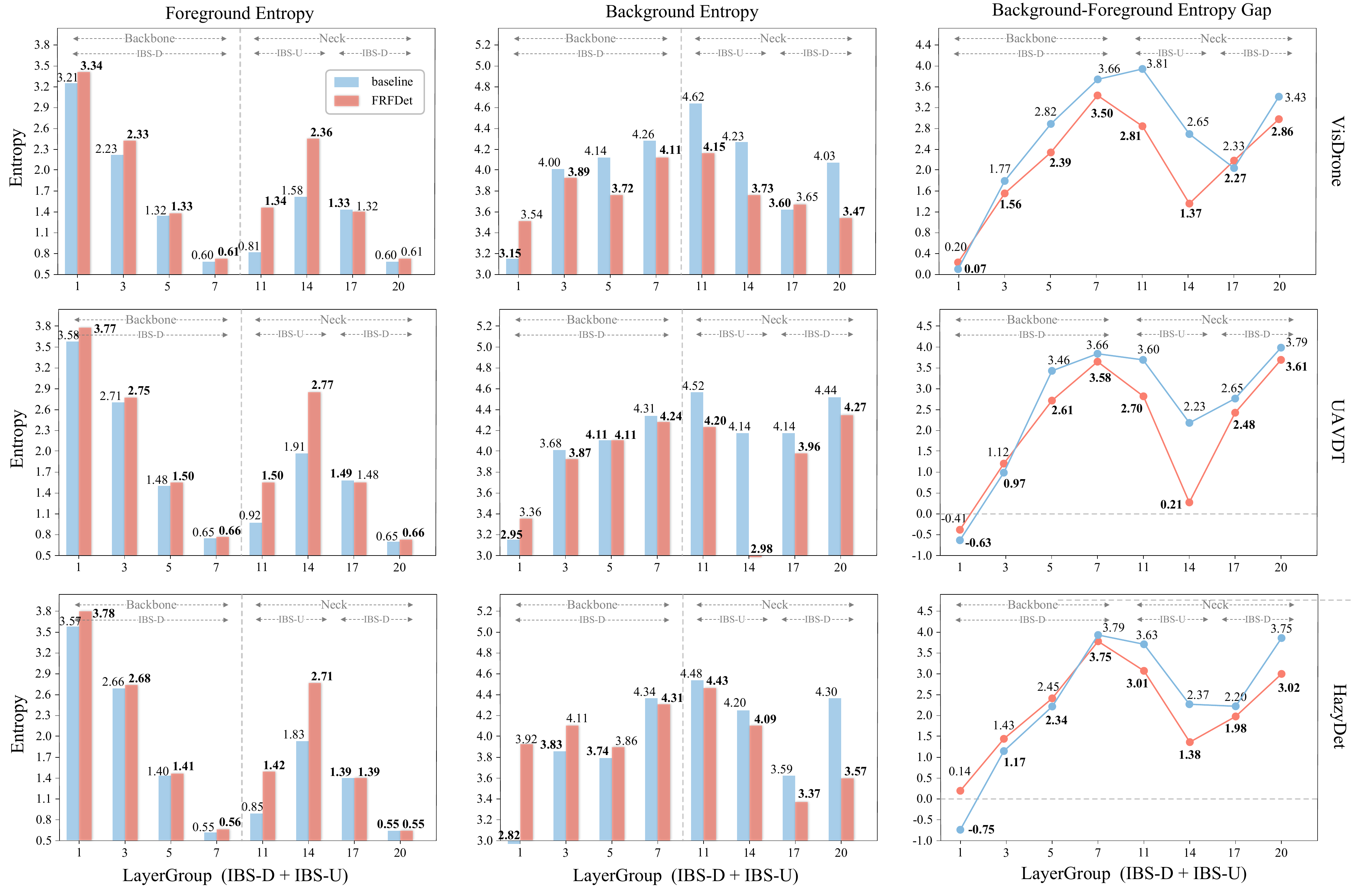}
    \caption{Quantitative visualization of foreground and background entropy for FRFDet’s IBS compared to the baseline model across three UAV datasets. Red and blue curves represent FRFDet and the baseline model, respectively. “LayerGroup (IBS-D+IBS-U)” denotes a set of non-contiguous layers that collectively constitute the IBS module. “Background-Foreground Entropy Gap” is defined as the background entropy minus the foreground entropy. Values in bold represent superior results.}
    \label{fig:ibs-entropy}
\end{figure*}

\begin{figure*}
    \centering
    \includegraphics[width=1.0\linewidth]{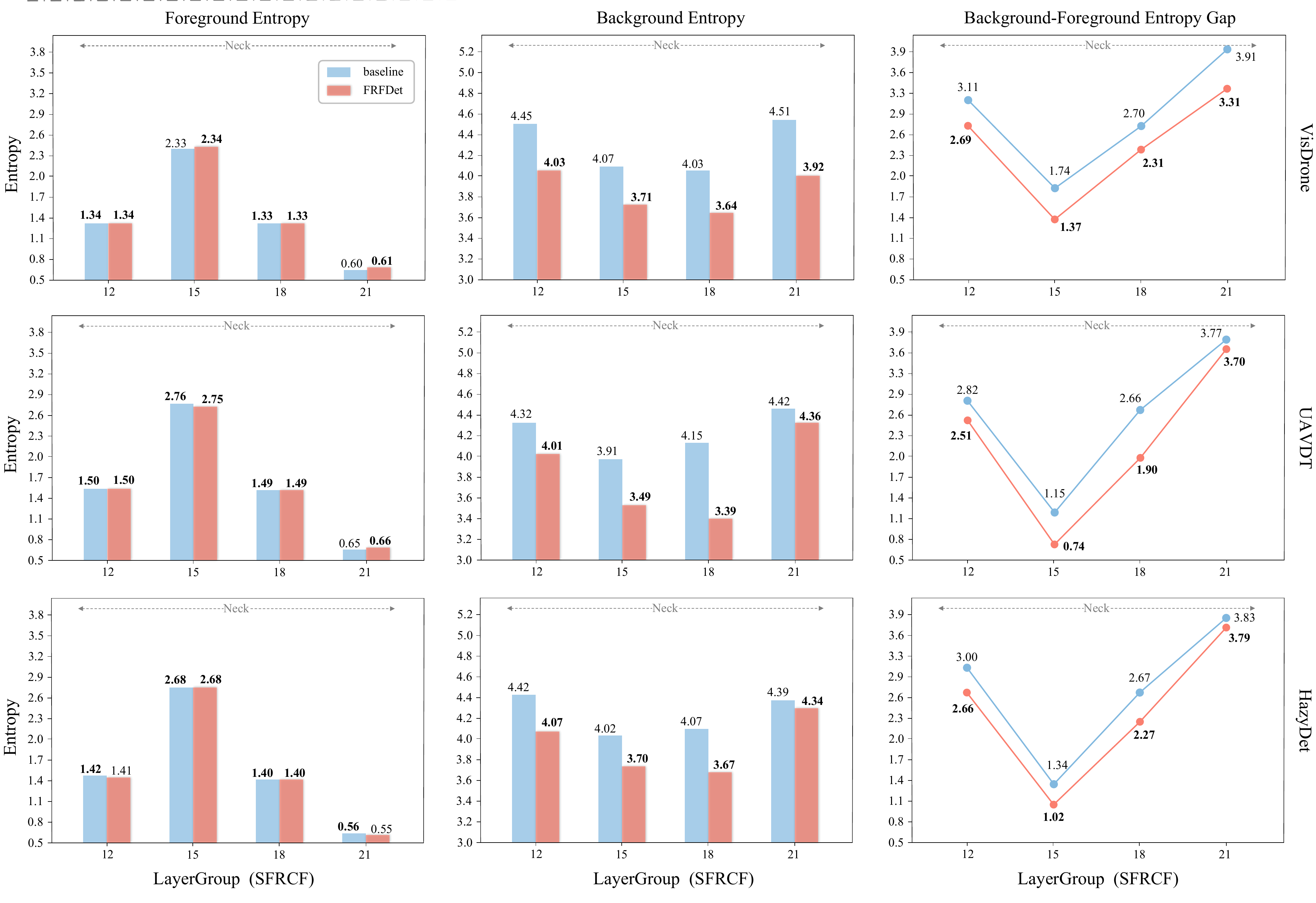}
    \caption{Quantitative visualization of foreground and background entropy for the SFRCF module in our FRFDet compared to the baseline model across three UAV datasets. Red and blue curves represent FRFDet and the baseline model, respectively. “LayerGroup (SFRCF)” denotes a set of non-contiguous layers that collectively constitute the SFRCF module. Values in bold represent superior results.}
    \label{fig:sfrcf-entropy}
\end{figure*}

\begin{figure}[t]
    \centering
    \includegraphics[width=0.98\linewidth]{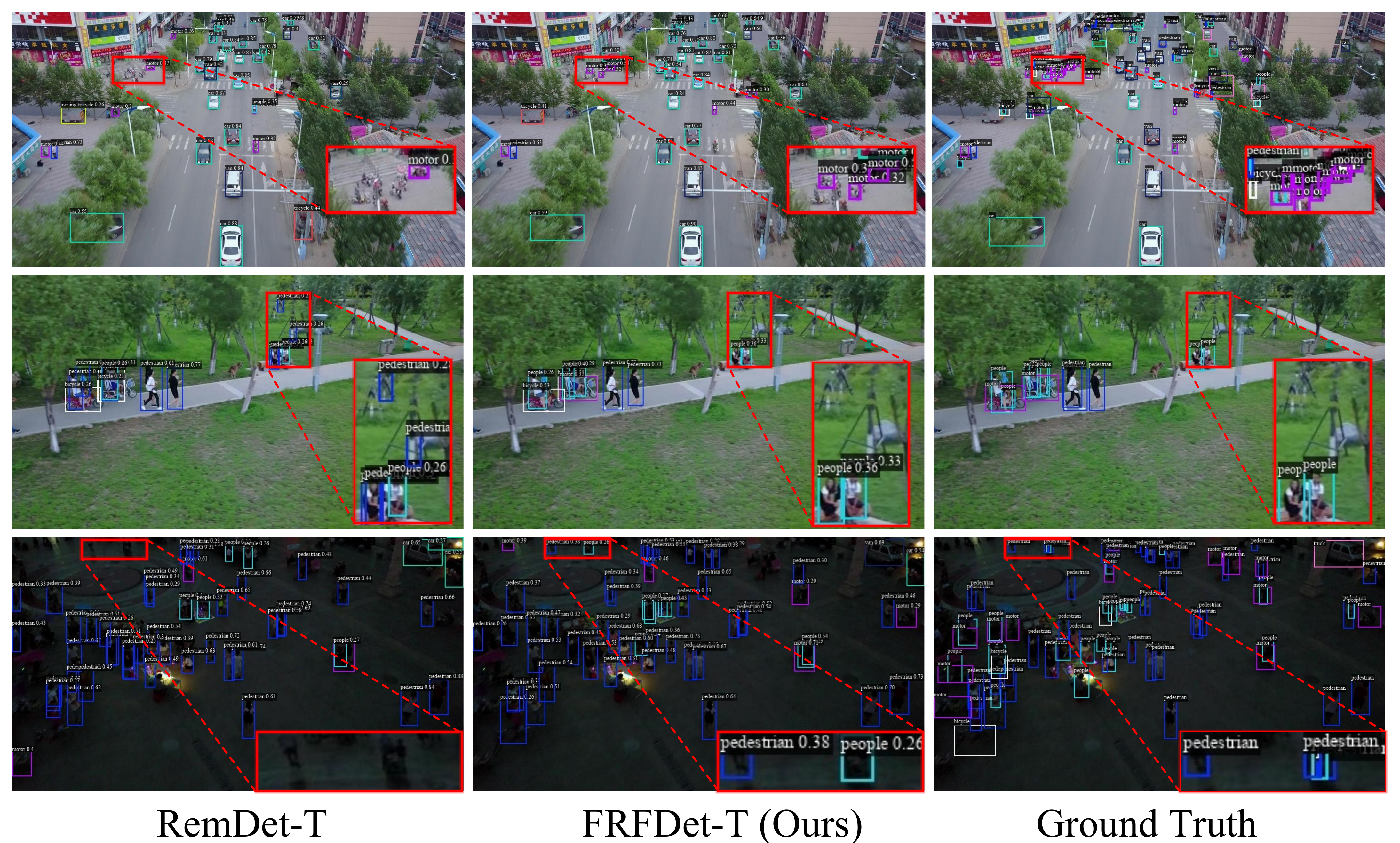}
    \caption{Visualization of detection results on VisDrone, comparing our FRFDet-T with RemDet-T under the from-scratch training mode.}
    \label{fig:det_res_visdrone}
\end{figure}

\begin{figure*}[t]
    \centering
    \includegraphics[width=1.0\textwidth]{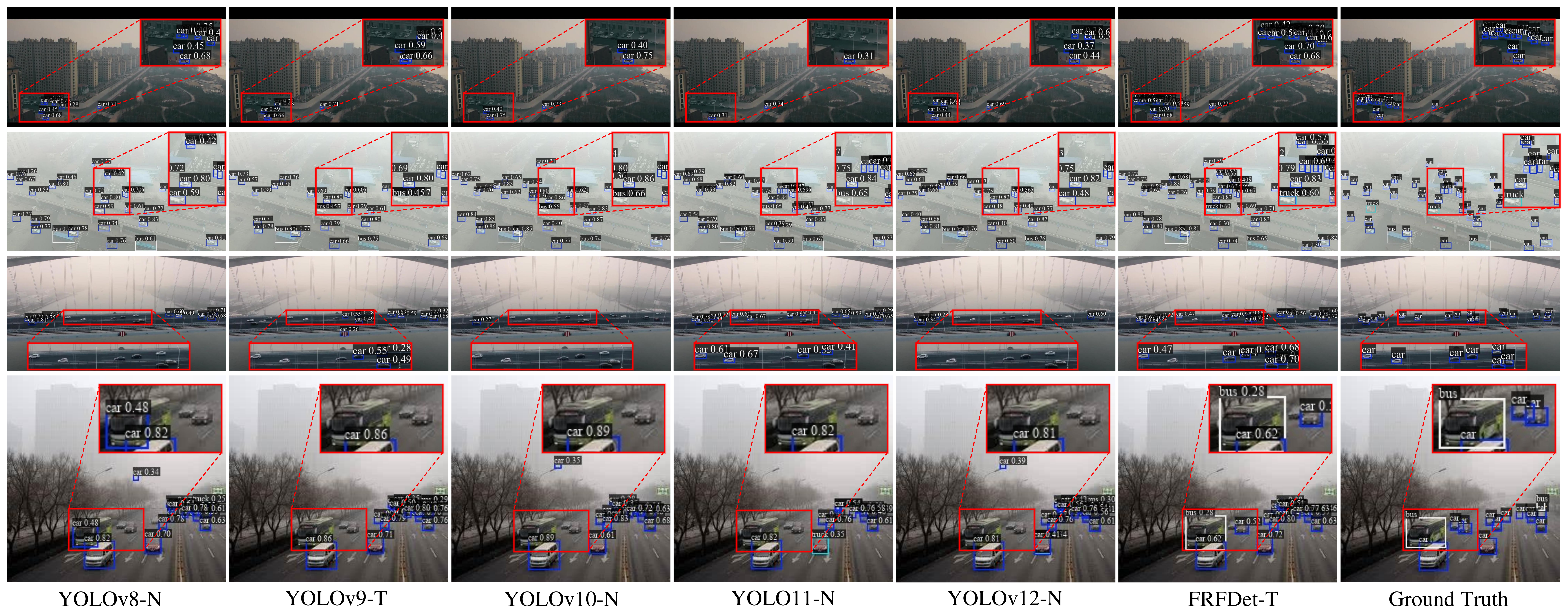}
    \caption{Visualization of detection results on HazyDet's RDDTS, comparing our FRFDet-T with state-of-the-art real-time detectors under the from-scratch training mode.}
    \label{fig:det_res_hazydet}
\end{figure*}

\section{Analysis of Background and Foreground Entropy}

In addition to the quantitative analysis (fig. 5) and the qualitative visualizations of feature maps (fig. 3 and fig. 4) and inter-channel similarity (SSIM) (fig. 3) provided in the main paper, we further investigate the effectiveness of our proposed components (IBS and SFRCF) from an information-theoretic perspective. Specifically, we analyze the entropy difference between foreground and background regions to reveal the key factors behind the effectiveness of our method. 

Given the ground-truth bounding boxes, we divide each feature map into foreground and background regions. For each region, we compute the Shannon entropy as follows:

Let a region contain a set of feature values:
\begin{gather}
    x=\left \{x_{1}, x_{2},..., x_{n}\right \}, x_{i}\in \mathbb{R}  
\end{gather}

1. Discretization into histogram bins:
Divide the value range of $x$ into $K$ equally spaced bins ${B_{1}, B_{2}, ..., B_{K}}$, and compute the bin counts $c$:
\begin{gather}
    c_{j} = \#\left \{x_{i} \in B_{j}\right \}, j = 1,2,...,K
\end{gather}

2. Probability normalization:
\begin{gather}
    p_{i} = \frac{c_{i}}{\sum_{j=1}^{K}c_{j} + \epsilon}, \epsilon > 0
\end{gather}

Here, we set $\epsilon$ to $1\times10^{-9}$.

3. Entropy computation:
\begin{gather}
    H(x) = -\sum_{i=1}^{K}p_{i} \cdot log_{2}(p_{i}), p_{i} > 0
\end{gather}

Building on the above formulation, we computed the average foreground and background entropy across layer groups (each comprising multiple target layers) for IBS and SFRCF on three UAV datasets, and tracked the evolution of the background-foreground entropy gap with increasing depth.

As illustrated in \cref{fig:ibs-entropy}, we present a quantitative comparison of foreground and background entropy for IBS across three UAV datasets. The first column reveals that IBS consistently increases foreground entropy compared to the baseline, with the most notable gains observed at layer groups 11 and 14. These groups correspond to the IBS-U module, characterized by a learnable symmetric structure. This suggests that IBS effectively captures richer discriminative features throughout the feature transmission process, while the symmetric sampling mechanism enhances the model's focus on target regions by improving foreground feature expressiveness. The second column shows that IBS significantly reduces background entropy, indicating a suppression of background redundancy and noise. This leads to more compact and informative representations, improving the quality of learned features. Consequently, the combination of increased foreground entropy and decreased background entropy results in a smaller background-foreground entropy gap, as depicted in the third column. This reduced gap implies enhanced foreground saliency, further validating the core strength of IBS—its ability to suppress background clutter while preserving and enhancing discriminative features that are critical for accurate small object detection. 

However, a noteworthy phenomenon is observed in the third row, corresponding to the HazyDet dataset—a representative foggy UAV scenario. During the feature extraction stage, our method exhibits higher background entropy than the baseline. This increase is attributed to the elevated scene complexity and intensified foreground-background ambiguity introduced by haze, which compels the model to attend to background regions that are visually entangled with the foreground in the early layers. As the network deepens, it gradually learns to disentangle and distinguish foreground from background. This progressive refinement is illustrated in the third column of the third row, where the gap between the red and blue curves consistently narrows from the 1st to the 5th layer group. From the 7th layer group onward, the entropy gap continues to decrease, reflecting enhanced discriminative capacity and more precise feature representation in deeper stages.

Similarly, we further analyze the SFRCF module in the Neck, as illustrated in \cref{fig:sfrcf-entropy}. Across all three UAV datasets, the foreground entropy of our method remains comparable to that of the baseline, indicating that SFRCF effectively preserves the discriminative capability for foreground objects. In contrast, the background entropy of our approach is significantly lower than that of the baseline, demonstrating that SFRCF offers substantial advantages in aligning multi-scale semantic features while suppressing uncertainty and noise in irrelevant background regions. Moreover, the reduced entropy gap between background and foreground further validates that our SFRCF module more effectively delineates the semantic boundaries between foreground and background throughout the fusion process.

\section{More Visualization Results}

\subsection{Qualitative Results under VisDrone}
As depicted in \cref{fig:det_res_visdrone}, we compare FRFDet-T with RemDet-T on the VisDrone dataset. FRFDet-T not only detects more small objects but also minimizes misclassifications between similar categories, such as people and pedestrians.

\subsection{Qualitative Results under Foggy UAV Conditions}
As shown in \cref{fig:det_res_hazydet}, FRFDet demonstrates robust detection under foggy conditions characterized by reduced visibility and significant noise. It accurately localizes targets, whereas other methods often produce missed detections or misclassifications (e.g., misclassifying a truck as a bus). These results highlight FRFDet’s resilience in challenging UAV scenarios and its strong generalization capability in real-world environments.

\section{Limitations and Future Directions}

Although FRFDet achieves state-of-the-art performance on representative UAV benchmarks and the large-scale MS COCO dataset, and demonstrates strong generalization in real-world foggy conditions, it still follows a supervised learning paradigm and relies on single-modality inputs. Consequently, its generalization capability in complex and unseen real-world scenarios remains limited. 

Recent advancements in Visual Language Large Models (VLLMs), such as Qwen3-VL \cite{Qwen3-VL}, GLM-4.1V \cite{GLM4_1V}, and PAM \cite{PAM}, have showcased remarkable capabilities in generalization, visual grounding, and instance-level perception, including tasks like visual grounding and instance segmentation. Meanwhile, diffusion models \cite{wang2026personalq} have emerged as a powerful paradigm for high-quality data generation, offering new opportunities for constructing large-scale annotated datasets to support the training of VLLMs. These models achieve robust performance by leveraging supervised fine-tuning (SFT) or reinforcement fine-tuning (RFT) paradigms with high-quality data and multimodal capabilities. Despite these achievements, applying VLLMs to small object detection in unmanned aerial vehicle (UAV) imagery remains an underexplored yet critical research direction. This task demands the development of high-quality, UAV-specific multimodal benchmarks to address challenges such as dense object distributions, high-resolution imagery, and hallucination caused by noisy backgrounds. By tackling these issues, multimodal methods can strive to achieve detection precision comparable to specialized visual small object detection models, thereby advancing the field of aerial perception.

In future work, we plan to explore the integration of UAV small object detection with language-image-video multimodal capabilities to enhance robustness and generalization across diverse UAV scenarios.

% \bibliographystyle{IEEEbib}
% \bibliography{icme2026references}

% \end{document}

\end{document}